\documentclass[11pt,a4paper]{article}
\usepackage[hyperref]{acl2017}
\usepackage{times}
\usepackage{latexsym}
\usepackage{graphicx}

\usepackage{amsmath}
\usepackage{microtype}
\usepackage{array}

\aclfinalcopy 

\newcolumntype{C}[1]{>{\centering\let\newline\\\arraybackslash\hspace{0pt}}m{#1}}

\title{Inferring Narrative Causality between Event Pairs in Films}

\author{Zhichao Hu and Marilyn A. Walker \\
  Natural Language and Dialogue Systems Lab \\
  Department of Computer Science, University of California Santa Cruz \\
  Santa Cruz, CA 95064, USA \\
  {\tt zhu@soe.ucsc.edu}, {\tt mawalker@ucsc.edu}\\}
\date{}

\begin{document}
\maketitle
\begin{abstract}

To understand narrative, humans draw inferences about the underlying
relations between narrative events.  Cognitive theories of narrative
understanding define these inferences as four different types of
causality, that include pairs of events A, B where A physically causes
B (X drop, X break), to pairs of events where A causes emotional state
B (Y saw X, Y felt fear). Previous work on learning narrative
relations from text has either focused on ``strict'' physical
causality, or has been vague about what relation is being learned.
This paper learns pairs of causal events from a corpus of film scene
descriptions which are action rich and tend to be told in
chronological order.  We show that event pairs induced using our
methods are of high quality and are judged to have a stronger causal
relation than event pairs from Rel-grams.

\end{abstract}

\section{Introduction}
\label{sec:intro}

Telling and understanding stories is a central part of human
experience, and many types of human communication involve narrative
structures. Theories of narrative posit that {\sc narrative causality}
underlies human understanding of a
narrative~\cite{warren1979event,trabasso1989logical,van1990causal}.
However previous computational work on narrative schemas, scripts or
event schemas learn ``collections of events that tend to co-occur''
\cite{ChambersJurafsky08,balasubramanian2013generating,PichottaMooney14},
rather than causal relations between events \cite{Rahimtoroghietal16}.
Another limitation of previous work is that it has mostly been applied
to newswire, limiting what is learned to relations between newsworthy
events, rather than everyday events
\cite{Rahimtoroghietal16,Huetal13,BeamerGirju09,Manshadietal08}.

Our focus here is on {\sc narrative causality}~\cite{trabasso1989logical,van1990causal}, the four different relations
posited by narrative theories to underly narrative coherence:

\begin{itemize}
\setlength\itemsep{0em}
	\item {\sc physical}: Event A physically causes event B to happen
	\item {\sc motivational}: Event A happens with B as a motivation
	\item {\sc psychological}: Event A brings about emotions (expressed in event B) 
	\item {\sc enabling}: Event A creates a state or condition for B to happen. A enables B. 
\end{itemize}

Previous work on learning causal relations has primarily focused on
physical causality~\cite{RiazGirju10,BeamerGirju09}, while our aim is
to learn event pairs manifesting all types of narrative causality,
and test their generality as a source of causal knowledge.  We
posit that film scene descriptions are a good resource for learning
narrative causality because they are: (1) action rich; (2) about
everyday events; and (3) told in temporal order, providing a primary
cue to causality~\cite{BeamerGirju09,Huetal13}. 

\begin{figure}[ht]
 \centering
\begin{tabular}{|c|p{2.6in}|}
\hline 
\bf \# & \bf Scene \\ \hline
1& Pippin, sitting at the bar, chatting with
Locals. Frodo leaps to his feet and pushes his way towards the
bar. Frodo \textbf{grabs} Pippin's sleeve, \textbf{spilling} his
beer. Pippin \textbf{pushes} Frodo away... he \textbf{stumbles}
backwards, and \textbf{falls} to the floor. \\ \hline
2 & Bilbo leads Gandalf into Bag
End... Cozy and cluttered with souvenirs of Bilbo's travels. Gandalf
has to \textbf{stoop} to \textbf{avoid} hitting his head on the low
ceiling. Bilbo hangs up Gandalf's hat on a peg and trots off down the
hall. Bilbo disappears into the kitchen as Gandalf \textbf{looks}
around.. \textbf{enjoying} the familiarity of Bag End... He
\textbf{turns}, \textbf{knocking} his head on the light and then
walking into the wooden beam. He groans.\\ 
\hline 
3 & Bilbo \textbf{pulls out} the ring... he \textbf{stares at} it in his palm. With all his will power, Bilbo \textbf{allows} the ring to slowly \textbf{slide off} his palm and drop to the floor. The tiny ring lands with a heavy thud on the wooden floor. \\
\hline 
4 & GANDALF... lying unconscious on a cold obsidian floor. He \textbf{wakes} to the sound of {ripping} and {tearing} ... \textbf{rising} onto his knees... lifting his head... Gandalf stands as the camera pulls back to reveal him stranded on the summit of Orthanc. \\

\hline 
   \end{tabular}
 \caption{Film Scenes from Lord of the Rings}
 \label{fig:lotr-film-scenes}
\end{figure}

Film scenes contain many descriptions encoding {\sc physical
  causality}, e.g. in Fig.~\ref{fig:lotr-film-scenes}, Scene 1, Frodo
grabs Pippin's sleeve, causing Pippin to spill his beer ({\it grab -
  spill}). Pippin then pushes Frodo away, causing Frodo to stumble
backwards and fall to the floor ({\it push - stumble}, {\it stumble -
  fall}, and {\it push - fall}).  But they also contain all other
types of narrative causality: in Scene 2, Gandalf has to
stoop, because he wants to avoid hitting his head on the low ceiling
({\it stoop - avoid}: {\sc motivational}). He then looks around, and
enjoys the result of looking: the familiarity of Bag End ({\it look -
  enjoy}: {\sc psychological}). He turns, which causes him to knock
his head on the light ({\it turn - knock}: the weak causality of {\sc
  enabling}).\footnote{Gandalf did not {\it turn} in order to {\it
    knock}, which would have been {\sc motivational}. Nor was it
  entailed that {\it turning} would cause {\it knocking}, which would
  have been {\sc physical}, because he clearly could have missed
  hitting his head if he had been more careful.}

This paper learns causal pairs from a corpus of 955 films.
 Because previous work
shows that more specific, detailed causal relations can be learned
from topic-sorted corpora \cite{RiazGirju10,Rahimtoroghietal16}, we
explore differences in learning between genres of film, positing e.g. that
horror films may feature very different types of events than comedies.
We also test the quality of what is learned when we train on genre specific
texts vs. the whole collection. Our results show that:
\begin{itemize}
\item  human judges
can distinguish between strong and weakly causal event pairs induced
using our method (Section~\ref{sec:exp1}); 
\item our strongly causal event pairs
are rated as more likely to be causal than those provided by the
Rel-gram corpus~\cite{balasubramanian2013generating} (Section~\ref{sec:exp2});  
\item human judges
can recognize different types of narrative causality (Section~\ref{sec:exp3}); 
\item using both
whole-corpus and genre-specific methods yields similar results for quality,
despite the smaller size of the genre-specific subcorpora. Moreover,
the genre-specific method learns some event pairs 
that are different than whole corpus event-pairs, while still being high-quality.
(Section~\ref{sec:exp4}); 
\end{itemize}

We explain our method in Section~\ref{sec:method}, and then present
experimental results in Section~\ref{sec:eval}.  We leave a more detailed discussion of
related work until Section~\ref{sec:related} when we can compare it
more directly with our own.

\section{Experimental Method}
\label{sec:method}

We estimate the likelihood of a narrative causality
relation between events in film scenes. 

\subsection{ Film Scenes \& Pre-Processing.} 
We chose 11 genres with more than 100 films from a corpus of film
scene descriptions \cite{Walkeretal12d,Huetal13},\footnote{From {\tt
https://nlds.soe.ucsc.edu/fc2}} resulting in 955 unique films. 
Film scripts were scraped from the IMSDb website, film 
dialogs and scene descriptions were then automatically separated. 
Films per genre range from 107 to 579. Films can
belong to multiple genres, e.g. the scenes from \emph{The Fellowship
  of the Ring} shown in Figure~\ref{fig:lotr-film-scenes} would become
part of the genres of Action, Adventure, and Fantasy.  Each film's 
scene descriptions ranges from 2000 to 35000 words.
Table~\ref{tab:film-stats} enumerates the sizes of each genre,
illustrating the potential tradeoff between getting good probability
estimates for event co-occurrence when the same events are repeated
{\bf within} a genre, vs. across the whole corpus.  We use Stanford
CoreNLP 3.5.2 to tokenize, lemmatize, POS tag, dependency parse and
label named entities \cite{manningetal14}.

\begin{table*}[hbt]
	\centering
\begin{tabular}{|l|c|r|p{2.5in}|}
	\hline
\bf Genre & \# \bf Films &\bf Word Count &\bf Example\\ \hline
Action&	290&3,758,387 &The Avengers\\
Adventure& 	166&2,115,247  &Indiana Jones and the Temple of Doom\\
Comedy& 	347&3,434,612 &All About Steve \\
Crime& 	201&2,342,324 &The Italian Job \\
Drama& 	579& 6,680,749&American Beauty \\
Fantasy& 	113&1,186,587 &Lord of the Rings: Fellowship of the Ring\\
Horror& 	149&1,789,667 &Scream \\
Mystery& 	107&1,346,496 &Black Swan \\
Romance& 	192&2,022,305 &Last Tango in Paris \\
Sci-Fi& 	155&1,964,856 &I, Robot \\
Thriller& 	373&4,548,043 &Ghost Rider \\ \hline
	\end{tabular}
	\caption{Distribution of Films By Genre.	\label{tab:film-stats}}
\end{table*}

\subsection{Compute Event Representations.}
An event is defined as a verb lemma, as in previous
work~\cite{ChambersJurafsky08,DoChaRo11,RiazGirju10,Manshadietal08}.
We extract events by keeping all tokens whose POS tags begin with {\tt
  VB}: {\tt VB}, {\tt VBD}, {\tt VBG}, {\tt VBN}, {\tt VBP}, and {\tt
  VBZ}. This results in extracting deverbal nouns that implicitly
evoke events, such as the events of {\it ripping} and {\it tearing} in
Scene 4 of Figure~\ref{fig:lotr-film-scenes}.  This definition also
allows us to pick up {\it resultative clauses} along with the 
action that caused the result
\cite{hovav2001event,goldberg2004english}, e.g. in {\it He slammed the
  door shut}, both {\it slammed} and {\it shut} are picked up as
verbs. We exclude light verbs e.g. \emph{be, let, do, begin, have,
  start, try}, because they often only represent a meaningful event
when combined with their complements.
 
We extract the subject (\emph{nsubj, agent}), direct object
(\emph{dobj, nsubjpass}), indirect object (\emph{iobj}) and particle
of the verb (\emph{compound:prt}), if any. In order to abstract and
merge different arguments, we generalize the arguments to two types:
\emph{person} and \emph{something}. We generalize an argument to
\emph{person} when: (1) its named entity type is {\sc person}; or (2) it is a
pronoun (except ``it''); or (3) it is a noun in WordNet with more than
half of its Synsets having lexical filename {\tt noun.person}, e.g. 
{\it doctor, soldier, waiter, man, woman}. Our narrative causal semantics would
be more specific if we could generalize over other types of named
entities as well, such as {\it location}. However 
Stanford NER identifiable named entities rarely occur
in film data.

\begin{figure*}[t!]
\centering
\begin{tabular}{|c|}
\hline
		\includegraphics[width=0.98\textwidth]{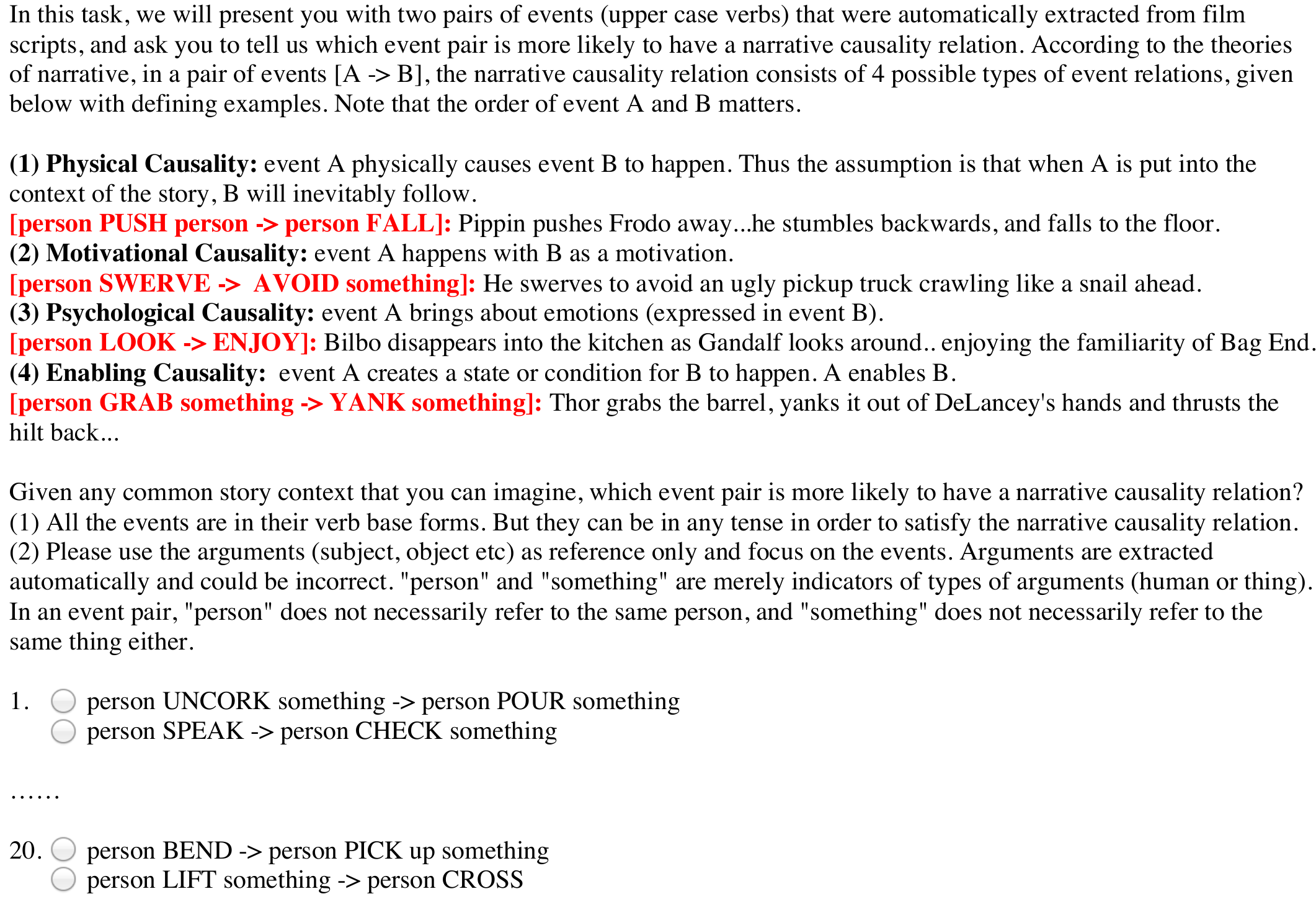}\\
\hline
\end{tabular}
	\caption{Instructions for the MT HIT. 		\label{fig:mturk_hit}}
\end{figure*}

For every event, we record the combinations of its arguments and
particle for every instance. For example, the instance of event
``pick'' in sentence: {\it He picked it up... a pearl}, has
combination \emph{subj: person, dobj: something, iobj: none, particle:
  up}. We pick the combination with the highest frequency to represent
the arguments and particle for each event. 

\subsection{Calculating Narrative Causality.}
We use the
Causal Potential (CP) measure in (1), shown to work well in previous work
\cite{BeamerGirju09,Huetal13}:

\begin{equation}
\mathit{CP}(e_1, e_2) = \mathit{PMI}(e_1, e_2) + \log\frac{P(e_1\rightarrow e_2)}{P(e_2\rightarrow e_1)}\\
\label{eq:cp}
\end{equation}
\begin{equation*}
\mathrm{where}~\mathit{PMI}(e_1, e_2) = \log\frac{P(e_1, e_2)}{P(e_1)P(e_2)}
\end{equation*}

where the arrow notation means ordered event pairs, i.e. event $e_1$
occurs before event $e_2$. CP consists of two terms: the first is
pair-wise mutual information (PMI) and the second is relative ordering
of bigrams. PMI measures how often events occur as a pair (without
considering their order); whereas relative ordering accounts for the order
of the event pairs because temporal order is one of the strongest cues
to causality \cite{BeamerGirju09,RiazGirju10}.

We obtain the frequency of every event and event pair for each
genre. Unseen event pairs are smoothed with frequency equal to 1. 
In this paper, the notion of window size indicates how many events 
after the current event are paired with the current event.
We use window sizes 1, 2 and 3, and calculate narrative causality for
each window size. In film scenes, events are very densely distributed,
(see Figure~\ref{fig:lotr-film-scenes}), thus related event pairs are
often adjacent to one another, but the discourse structure of film
scenes, not surprisingly, also contain related events separated by
other events \cite{GS86,MannThom87}. For example, in Scene 3
of Figure~\ref{fig:lotr-film-scenes}, Bilbo pulling out the ring enables 
him to slide it off his palm later (\emph{pull out} - \emph{slide off}). 
Moreover, while related events
are less frequently separated (window size 3), we assume that
unrelated events will be filtered out by their low probabilities. We
thus define a \emph{CPC} measure, shown in (2) that combines the
frequencies across window size:

\begin{equation}
\mathit{CPC}(e_1, e_2) = \sum\limits_{i=1}^{w_{max}}\frac{\mathit{CP}_i(e_1, e_2)}{i}\\
\label{eq:combined-cp}
\end{equation}

where $w_{max}$ is the max window size. $\mathit{CP}_i(e_1,
e_2)$ is the CP score for event pair $e_1, e_2$ calculated using
window size $i$. The \emph{CPC} measure combines frequencies across
window sizes, but punishes event pairs from larger window sizes, thus
assuming that nearby events are more likely to be causal.

\section{Evaluation and Results}
\label{sec:eval}

We posit that human judgments are the best way to evaluate the quality
of the induced event pairs, as opposed to automatic measures such as
Narrative Cloze, which assume that the event pairs in a particular
instance of text can be used as held-out test data
\cite{ChambersJurafsky08}. Our first experiment tests whether event pairs
with high \emph{CPC} scores are more likely to have a
narrative causality relation. Our second experiment compares 
pairs with high \emph{CPC} scores with 
their corresponding top Rel-gram pairs. Our
third experiment tests whether annotators can distinguish narrative
causality types. Our final experiment compares the quality and type
of causal pairs learned on a per genre basis, vs. those learned
on the whole film corpus. 
	
\begin{table*}[ht!]
\centering
	\begin{tabular}{|c|p{2.7in}|p{2.7in}|}
	\hline
	\bf \# & \bf High CPC Pair & \bf Low CPC Pair \\
	\hline \hline 
	1 &  {[person]} \emph{clink} [smth] - [person] \emph{drink} [smth] 
           & [person] \emph{strike} - [person] \emph{give} [person] [smth] \\ \hline
	2 &  {[person]} \emph{beckon} - [person] \emph{come} 
          &  [smth] \emph{become} - [person] \emph{hide} \\ \hline
	3 &  {[person]} \emph{bend} - [person] \emph{pick} up [smth] 
          &  [person] \emph{lift} [smth] - [person] \emph{cross} \\ \hline
	4 & {[person]} \emph{cough} - [person] \emph{splutter} 
          &  [person] \emph{force} - [smth] \emph{show} [smth] \\ \hline
	5 &  {[person]} \emph{crane} - [person] \emph{see} [smth] 
          &  [person] \emph{fade} - [person] \emph{allow} [person] \\ 
	\hline
	\end{tabular}
	\caption{\label{tab:eval-results-high-low-pairs} Narratively Causal Pairs where all 5 annotators selected the High CPC pair.}
\end{table*}

\subsection{High vs. Low CPC Event Pairs}
\label{sec:exp1}

After processing all the data, we have a list of event pairs scored by
\emph{CPC}, and rank-ordered within each genre.  Some of the genre specific event
pairs seem to intuitively reflect their genre, however there are many
learned pairs that are in overlap across genres.
We select the top 3000 event pairs with high scores from all the
genres (``high pairs''). The number of event pairs from a genre is
proportional to the number of films in that genre. We also select the
bottom 6000 event pairs with low scores from all the genres using
similar method (``low pairs''). Since many pairs are duplicated across
genre, the high pairs and low pairs are then de-duplicated (two event
pairs are defined as equal if they have the same verbs in the same
order). We keep the arguments with the highest frequencies. This
result in 960 high pairs. If an event has no subject, ``person'' is
added as subject, since most events have human agents.

For every event pair in the 960 high pairs, we randomly select a low
pair in order to collect human judgments on Mechanical Turk.  The task
first introduces event and event pair definitions, then defines the
four types of narrative causality with corresponding examples. Turkers
are asked to select the event pair that is more likely to manifest a
narrative causality relation. Each HIT consists of 20 judgements, and
we collect 5 judgements per HIT. Because this task requires some care,
Turkers had to be prequalified. The qualification test aims to test Turkers'
understanding of narrative causality. It is similar to the task
itself, but with more obvious choices, such as high CPC pair \textit{open - reveal} 
vs low CPC pair \textit{pay - fade}.
Figure~\ref{fig:mturk_hit} shows a 
simplified version of the HIT instructions.\footnote{The full instructions
provide more examples and background information.}

\begin{table}[hbt]	
\centering	
\begin{tabular}{|c|c|c|}
	\hline
\bf Genre &\bf  \# High Pairs  &  \bf \% Causality \\ \hline
Action& 320  & 86.3\\ 
Adventure& 171 &86.6\\ 
Comedy& 384  &84.9\\ 
Crime& 23 &84.9\\ 
Drama& \bf{665}	& \bf{82.6}\\ 
Fantasy& \bf{127} & \bf{90.7}\\ 
Horror& 156	&  87.2\\ 
Mystery& 	122  &87.7\\ 
Romance& 215	 & 86.0 \\ 
Sci-Fi& 158	& 88.0\\ 
Thriller& 405 & 87.7 \\ \hline 
\end{tabular}
\caption{Percentages of high pairs that receive majority vote results by genre. \label{tab:genre-results}}
\end{table}



The results show that humans judge
the high pairs as more likely to have a narrative causality relation
in 82.8\% of items. Among those, all the items receive 3 or more 
votes for the high pairs. Overall, all five Turkers select the high CPC
pairs in 51\% of the items. The average pairwise Krippendorff's
Alpha score is respectable at 0.56.

Table~\ref{tab:eval-results-high-low-pairs} shows items where all 5
Turkers selected the high pair.  For example, \emph{clink - drink} in
Row 1 could have either a {\sc motivational} or {\sc enabling}
narrative causality depending on the context, but the causal relation
in either case is much clearer than with the low CPC pair \emph{strike
  - give}. Row 2 and Row 5 \emph{beckon - come} and \emph{crane -see}
both have {\sc enabling} causality which is a weakly causal relation,
but again more meaningful than their low CPC counterparts. In Row 3,
it is clear that a 
person often \emph{bends} with the motivation to \emph{pick up} something.
In row 4 a person \emph{cough}s, {\sc physically} causes him to \emph{splutter}
everywhere.


\begin{table*}[hbt!]
\begin{small}
\centering
\begin{tabular}{|c|p{2.2in}|p{3in}| C{0.4in}|}
\hline \bf\# & \bf Narrative Causality (CPC) Pairs &\bf Rel-gram Pairs & \bf CPC Vote \#\\
 \hline  1 &{[person]} \emph{clear} [smth] - [person] \emph{reveal} [smth] & [person] \emph{clear} [smth] - [person] \emph{hit} [smth] & 5\\ 
2 &{[person]} \emph{embrace} - [person] \emph{kiss} & [person] \emph{embrace} [person] - [person] \emph{meet}   [person] & 5\\
3 &{[person]} \emph{empty} [something] -[person] \emph{reload} & [person] \emph{empty} [smth] - [person] \emph{shoot} [person] & 5 \\
4 &{[person]} \emph{marry} [person] - [person] \emph{think} & [person] \emph{marry} [person] - [person] \emph{die} [something] & 5 \\ 
5 &{[person]} \emph{stumble} - [smth] \emph{fall} & [person] \emph{stumble}
     upon [person] - [person] \emph{take} [person] & 5\\ \hline
6    &{[person]} \emph{gaze} - [smth] \emph{drift} & [person]
     \emph{gaze} at [person]- [person] \emph{see} [person] & 0\\
7   &{[person]} \emph{reveal} [smth] - [person] \emph{sit} & [person]
     \emph{reveal} [person] - [person] \emph{see} [person] & 0\\
8     &{[person]} \emph{watch} - [person] \emph{appal} & [person]
     \emph{watch} [person] - [person] \emph{see} [person] & 0\\
     \hline
\end{tabular}
\end{small}
\caption{Items where either CPC event pairs or Rel-gram event pairs were strongly
  preferred. 	\label{tab:eval-results-nc-relgram-pairs}}
\end{table*}

Table~\ref{tab:genre-results} shows majority vote results for
percentages of high pairs that are considered to exhibit more
narrative causality, sorted by genre. The results for all genres are
good, ranging from $\sim$82\% to $\sim$91\%.  Interestingly, Drama has
the highest number of films with the lowest percentage of judged
narrative causality, while Fantasy has the lowest number of films with
the highest judged narrative causality. This may be because the Drama
category is a catch-all (over half of the films are categorized this
way suggesting that it has low coherence as a genre). The poor
performance on Drama would then be consistent with previous work that
shows that topical coherence (genre in this case) improves causal
relation learning \cite{Rahimtoroghietal16,RiazGirju10}. We will return to
this point in Section ~\ref{sec:exp4}.

\subsection{CPC vs. Rel-gram Event Pairs}
\label{sec:exp2}

We then compare the narrative causality event pairs (high pairs) with
event pairs from the Rel-grams corpus
\cite{balasubramanian2012rel,balasubramanian2013generating}. 
Rel-grams (Relational n-grams) are pairs of open-domain relational 
tuples (T,T'). They are analogous to lexical n-grams,  but is computed 
over relations rather than over words.
For example, ``A person who gets arrested is typically charged 
with some activity.'' yield the tuple: 
T = ([police] \textit{arrest} [person]) and 
T' = ([person] \textit{be charge with} [activity]).
Over 1.8M news wire documents are used to build a database
of Rel-grams co-occurence statistics.

Using a
similar HIT template, we randomly sample 100 high CPC event pairs from
the 960 high CPC pairs, where we ensure that each of the first events
of the pairs are distinct. We use the publicly available search
interface for Rel-grams\footnote{http://relgrams.cs.stonybrook.edu/} to
find Rel-gram statement pairs that have the same first event. Modeling
our own experimental setup we set the co-occurrence
window to 5\footnote{The search interface does not support a window 
size of 3, thus we chose 5 as it's the closest window size larger 
than 1.}, and select the Rel-gram  pair with the highest
\#50(FS) (frequency of first statement occurring before second
statement within a window of 50).


To make Rel-gram event pairs similar to ours, we generalize their
arguments to ``person'' and ``something'' manually. We keep the verb
particle if any. For example, the Rel-gram pair ``[person]
\emph{remain} in [location] - [person] \emph{become} [leader]'' is
generalized to ``[person] \emph{remain} in [something] - [person]
\emph{become} [something]''. It is possible that this disadvantages
Rel-grams in some way, but our main focus is on the causality relation
between verbs, which should not be affected.
Moreover the two sets of event pairs cannot be compared
without this generalization. The same 5 annotators participate in this
5 HITs (100 items).



The results show that humans judge the CPC pairs to be more likely to
manifest a narrative causality relation 81\% of the time.  The average
pairwise Krippendorff's Alpha score of all Turkers is
0.482. Table~\ref{tab:eval-results-nc-relgram-pairs} shows items where
all Turkers judge the CPC pairs as more likely to be causally related.
For example, in Row 1 to \emph{clear} seems more likely to enable
something being \emph{revealed}, instead of causing a person to
\emph{hit} something.  In Row 2, even though \emph{embrace} and
\emph{kiss} might only have an {\sc enabling} narrative causality
relation, the reversed causality between \emph{embrace} and
\emph{meet} in the Rel-gram pair is based on symmetric conditional
probability (SCP) rather than explicit causal modeling. SCP combines Bigram
probability in both directions as follows:
\begin{equation}
SCP(e_1, e_2) = P(e_2|e_1) \times P(e_1|e_2)
\end{equation}

In Row 4, \emph{marrying}
someone might just possibly enable one to think about something, but could hardly
enable/cause someone to die. In Row 5  \emph{stumble} physically causes one to
\emph{fall}, while it is more difficult to see the causal relation
between \emph{stumbling on} someone and then a person \emph{taking} another
person (somewhere).


\begin{table}[h]
\centering
\begin{tabular}{|C{1.2in}|c|c|}
\hline
\bf Narrative Causality Type & \bf Count  & \bf Example Pair\\ \hline
Physical & 13 & \emph{fire - blast}\\
Motivational & 29 &\emph{bend - retrieve} \\
Psychological & 9 &\emph{look - astonish}\\
Enabling & 28 & \emph{lean - whisper} \\ \hline
\end{tabular}	
\caption{Distribution of narrative causality types \label{tab:type-result-counts}.}
\end{table}

\begin{table*}[ht!]
\small
\centering
	\begin{tabular}{|l|c|l|c|}
	\hline
	\bf Fantasy & \bf CPC & \bf Action  & \bf CPC \\
	\hline 
	{[person]} \emph{slam} [smth] - \emph{shut} & 4.95 & [person] \emph{huff} - [person] \emph{puff} & 5.57 \\
	\emph{send} [smth] - [smth] \emph{fly}& 4.89 & \emph{bind} - \emph{gag} & 5.50 \\
	{[person]} \emph{watch} - [smth] \emph{disappear} & 4.87 & [smth] \emph{swerve} - \emph{avoid} [smth] & 5.21 \\
	{[person]} \emph{turn} - \emph{face} [person] & 4.83 & [person] \emph{bend} - [person] \emph{pick} up [smth] & 5.01\\
	{[person]} \emph{pull} [smth] - \emph{reveal} [smth] & 4.70 & \emph{send} [smth] - [smth] \emph{tumble} & 4.85\\
	{[person]} \emph{pick} up [smth] - \emph{carry} [smth] & 4.54 & \emph{send} [smth] - \emph{sprawl}& 4.83\\ 
	{[person]} \emph{reach} - [person] \emph{pull} [smth] & 4.42 & {[person]} \emph{slam} [smth] - \emph{shut} & 4.79\\ 
	\hline
	\bf Sci-Fi & \bf CPC & \bf Thriller   & \bf CPC \\
\hline	
	{[person]} \emph{bend} - [person] \emph{pick} up [smth] & 4.88  &  \emph{bind} - \emph{gag} & 5.66 \\
	\emph{follow} - [person] \emph{gaze} & 4.83 &  [smth] \emph{swerve} - \emph{avoid} [smth] & 5.37 \\
	{[person]} \emph{grab} [smth] -  [person] \emph{yank} [smth] & 4.83 & [person] \emph{rummage} - [person] \emph{find} [smth]& 5.05 \\
	\emph{send} [smth] - [smth]  \emph{fly} & 4.81 & {[person]} \emph{inhale} - peroson \emph{exhale} &  5.04\\
	{[person]} \emph{slam} [smth] - \emph{shut} & 4.78 & {[person]} \emph{slam} [smth] - \emph{shut} & 5.00 \\
	{[person]} \emph{grab} [smth] - [person] \emph{drag} [person] & 4.77 & \emph{send} [smth] - [smth]  \emph{fly} &  4.97\\
	{[person]} \emph{reach} - \emph{touch} [smth] & 4.67 & [person] \emph{reach} - [person] \emph{produce} [smth]&  4.81\\
	\hline
	\end{tabular}
	\caption{Event pairs with Highest CPC scores from Fantasy, Action, Sci-Fi and Thriller genres.}
		\label{tab:good-genre-pairs}
\end{table*}

\subsection{Narrative Causality Types}
\label{sec:exp3}

Although theories of narrative posit four different types of narrative
causality, previous work has not conducted reliability studies with
non-experts such as Turkers.  Here we explore whether humans can
distinguish narrative causality types, by asking Turkers to decide
which relation holds between an event pair. The instructions contain
descriptions of narrative causality types and the strength of these
relations (from strong to weak: {\sc physical}, {\sc motivational},
{\sc psychological} and {\sc
  enabling}~\cite{trabasso1989logical}). Because the stronger types of
narrative causality could also be considered {\sc enabling}, Turkers
are instructed to choose the strongest narrative causality that could
be applied to the event pair.

We select 100 pairs randomly from the high CPC pairs of the 479
questions that had the highest Turker agreement.
Among all 100 questions, 79\% of the items receive a majority vote
result (3 or more Turkers selecting the same answer).
The distribution of narrative causality types of the 79 items is shown
in Table~\ref{tab:type-result-counts}. Interestingly, films are full
of motivational causality, which often reflect action sequences where
protagonist pursue particular narratively relevant goals
\cite{RappGerrig06,RappGerrig02}.

\begin{table*}[hbt!]
\small
\centering
	\begin{tabular}{|l|p{5.5in}|}
	\hline
	\bf{Genre} & \bf{Event pairs}\\ \hline
Fantasy & \emph{struggle - get,  reveal - stand, see - stand, get - marry, sit - sip, nod - head, make - break, spin - face, take - bite, watch - disappear, pick - carry}\\ \hline

Sci-Fi &\emph{hear - echo, see - come, look - alarm, widen - see, head - stop, clear - reveal, sit - study, look - puzzle, peek - see}\\ \hline

Horror& \emph{listen - hear, stare - fascinate, hear - muffle, slow - stop, peel - reveal, reach - yank, reach - handle, grab - handle}\\ \hline

Mystery& \emph{slip - fall, dig - pull, walk - reach, look - confuse, sit - eat, knock - open, look - horrify, stop - look, sit - look, seem - lose}\\ \hline

Thriller& \emph{look - wonder, raise - fire, poise - strike, sit - hunch, rape - murder}\\ \hline

All& \emph{sit - leg, whoop - holler, huff - puff, disappear - reappear, cease - exist, dive - swim, spur - gallop, offer - decline, contain - omit, hoot - holler, pay - heed}\\
	\hline
	\end{tabular}
	\caption{Event pairs unique to Fantasy, Sci-Fi, Horror, Mystery, Thriller genres and all films.}
		\label{tab:pairs-unique-to-genre}
\end{table*}

\subsection{Genre Specific Causality}
\label{sec:exp4}

Previous work suggests that topical coherence and similarity of events
within the corpus used for learning causal/contingent event relations
might be as important as the size of the corpus
\cite{RiazGirju10,Rahimtoroghietal16}. In other words, smaller corpora
filtered by topic or genre might be more useful than large undifferentiated
sets \cite{Riloff96}, although obviously very large corpora that are
topic or genre sorted could be even more useful. We therefore test
whether separating films by genre yields higher quality event pairs
than a method that combines all films, irrespective of genre. We assume
that the very notion of a film genre defines a set of films with similar
types of events. 

We first compute a list of \emph{CPC} scores using films from all
genres and take 960 event pairs with highest scores. Comparing the 960
event pairs from all films with the 960 pairs from merging genres
described in Section \ref{sec:exp1}, we find that 728 pairs overlap between
the two sets. Thus with the smaller genre-specific corpora we learn
more than 70\% of the same causal pairs. The results shown in 
Table~\ref{tab:genre-results}
suggest furthermore that the genre-specific pairs are high quality.

However, it is still possible that the 232 pairs from each set that
are not in overlap vary in quality from the 728 pairs that are in
overlap. We therefore pick 100 random pairs from each set, match the
pairs randomly to form items, and repeat the event pairs comparison
HIT with these pairs. The results suggest that there are no
differences between the two methods as far as quality: in 48
of the 100 questions, pairs from genre-separated method have
Turkers' majority vote, vs. in 52 of the 100 questions pairs from
combined genres have the majority vote. 

Moreover we obtain {\bf more} high-quality, reliable narrative causality
relations using both methods, and we learn some genre-specific causal
relations that we do not learn on the whole corpus.
Table~\ref{genre-overlap} shows the the overlap in learned pairs
amongst the top 30 CPC pairs in five of the most distinct genres
(genres with highest percentages in Table~\ref{tab:genre-results}:
Fantasy, Sci-Fi, Horror, Mystery and Thriller) vs. all films (All).
Mystery has the smallest overlap with “All”, followed by Fantasy and
Sci-Fi. 

\begin{table}[ht]
\centering
\begin{tabular}{l|ccccc}
\hline
\bf Genre & \bf All & \bf Thr & \bf Mys & \bf Hor & \bf Sci\\
\hline
Fan	&8	&9	&13	&15	&14\\
Sci	&8	&12	&14	&18	&\\
Hor	&10 	&14	&14	&		&\\
Mys	&7	&12	&		&		&\\
Thr	&18	&		&		&		&\\
\end{tabular}
\caption{Overlap in learned pairs among the most distinct genres (Fantasy, Sci-Fi, Horror, Mystery and Thriller) vs. all films (All). \label{genre-overlap}}
\end{table}

To illustrate some of the differences,  Table~\ref{tab:good-genre-pairs} shows
event pairs with the highest \emph{CPC} scores in Fantasy, Action, Sci-Fi and
Thriller genres. Table~\ref{tab:pairs-unique-to-genre} shows event pairs unique
to each genre within its top 30 CPC pairs.


We also compare our 960 pairs from merging genres described in 
Section \ref{sec:exp1} with 200 event pairs extracted from camping and storm personal 
blog stories in~\newcite{Rahimtoroghietal16}. The only pairs that overlap
are: \emph{sit} - \emph{eat}, \emph{play} - \emph{sing}, illustrating again that causal relations learned are not as dependent on the size of the corpus, as they are on its topical and event-based coherence. Since most previous work on narrative schemas, scripts, event schemas or rel-grams has only been applied to one large corpus of newswire (Gigaword corpus), these methods have only learned relations about newsworthy topics, and even then,
perhaps only the most frequent, highly common news events. In contrast, both
our approach and that of \newcite{Rahimtoroghietal16} learn fine-grained causal relations
that underly narratives, which we believe are more in the spirit of Schank's original
motivation for scripts \cite{Lehnert81,Schank77,wilensky82,Dejong79}.

\section{Related Work}
\label{sec:related}

\newcite{Huetal13} tested four methods for inducing pairs of adjacent
events with contingency/causality relations from film scenes,
including Causal Potential, Pointwise Mutual Information, Bigram Model
and Protagonist-based Model.  \newcite{Rahimtoroghietal16} also
used a modified version of the the CP measure, adjusted to account for
the discourse structure of personal narratives in blogs.  Here we use
a much larger set of films and apply different techniques and a
detailed evaluation. Our learned causal pairs 
and supporting film data are available for download~\footnote{https://nlds.soe.ucsc.edu/narrativecausality}.

\newcite{DoChaRo11} used a minimally supervised approach, based on
focused distributional similarity methods and discourse connectives,
to identify causality relations between events in PDTB in context (both verbs
and nouns) \cite{Prasadetal08}. They present a detailed formula for calculating contingency/causality
that takes into account several different kinds of argument overlap
between adjacent events. However they do not provide any evidence that
all the components of this formula actually contribute to their
results.

\newcite{Gordonetal11} used event ngrams and discourse cues to learn
causal relations from first person stories posted on weblogs and
evaluated them with respect to the COPA SEM-EVAL task.  Other related
work learns likely sequences of temporally ordered events but does not
explicitly model {\sc causality}
~\cite{ChambersJurafsky09,balasubramanian2013generating,Manshadietal08}.

Work on VerbOcean
\cite{chklovski2004verbocean} use lexical patterns to learn semantic
verb relations of similarity, strength, antonymy, enablement and
happens-before relations.  \newcite{balasubramanian2013generating} use
symmetric probability to learn semantically typed relational triples 
(actor, relation, actor), which
they call Rel-grams (relational n-grams), and
show that their schemas outperform previous work
\cite{ChambersJurafsky09}.  We thus compared our event pairs with
Rel-grams, showing that humans are more likely to perceive narrative causality
in our event pairs.

\section{Discussion and Future Work}

We present an unsupervised model based on Causal
Potential~\cite{BeamerGirju09} to induce event pairs with narrative
causality relations from film scenes in 11 genres.  Results from four
human evaluations show that narrative causality event pairs induced
using our method are of high quality, and are perceived as more
causally related than corresponding Rel-grams. We show that
humans can identify different types of narrative causality, but we
leave automatic identification of these to future work. We also show 
that inducing narrative causality event pairs using both whole-corpus 
and genre-specific methods yields similar results for quality,
despite the smaller size of the genre-specific subcorpora. Moreover,
the genre-specific method learns high quality event pairs that are different 
than whole corpus event-pairs.

We are looking into applying and evaluating our CPC method to other genre and topic sorted datasets such as books and personal blogs. We want to expand our set of event pairs with narrative causality relations, which could potentially aid text understanding, information extraction, question answering, and content summarization. We also aim to explore features for narrative causality type classification. Information such as event A physically causes event B, or event C enables event D could further help aforementioned applications.

\section*{Acknowledgements}

This research was supported by Nuance Foundation Grant SC-14-74, NSF \#IIS-1302668-002, NSF CISE CreativeIT \#IIS-1002921 and NSF CISE RI EAGER \#IIS-1044693.

\bibliographystyle{acl_natbib}

\end{document}